\documentclass[letterpaper]{article} 
\usepackage{aaai2026}  
\usepackage{times}  
\usepackage{helvet}  
\usepackage{courier}  
\usepackage[hyphens]{url}  
\usepackage{graphicx} 
\usepackage{booktabs}
\usepackage{multirow}
\usepackage{array}
\usepackage{amsmath}
\usepackage{amssymb}
\usepackage{tabularx}
\usepackage{afterpage}
\usepackage{subcaption}
\usepackage{natbib}  
\PassOptionsToPackage{round}{natbib}
\usepackage{caption} 
\usepackage{algorithm}
\usepackage{algorithmic}

\usepackage{xurl}
\usepackage{xcolor} 
\usepackage[hidelinks,hypertexnames=false]{hyperref}
\usepackage{bookmark} 

\usepackage[toc,page,header]{appendix}
\usepackage{minitoc}

\renewcommand \partname{}

\usepackage{amsmath,amsfonts,bm}

\def\eqref#1{equation~\ref{#1}}

\def\1{\bm{1}}

\DeclareMathAlphabet{\mathsfit}{\encodingdefault}{\sfdefault}{m}{sl}
\SetMathAlphabet{\mathsfit}{bold}{\encodingdefault}{\sfdefault}{bx}{n}

\newcommand\blfootnote[1]{%
  \begingroup
  \renewcommand\thefootnote{}\footnote{#1}%
  \addtocounter{footnote}{-1}%
  \endgroup
}

\title{WorldAgen: Unified State-Action Prediction with \\ Test-Time World Model Training}
\author{
    Chi Wan\textsuperscript{\rm 1}\textsuperscript{*},
    Kangrui Wang\textsuperscript{\rm 1}\textsuperscript{*\dag},
    Yuan Si\textsuperscript{\rm 1},
    Pingyue Zhang\textsuperscript{\rm 1},
    Manling Li\textsuperscript{\rm 1}
}
\affiliations{
    \textsuperscript{\rm 1}Northwestern University \\[4pt]
    \href{https://worldagen.github.io}{\textcolor[rgb]{0.42,0.50,0.84}{\textbf{https://worldagen.github.io}}}
}

\begin{document}

\maketitle

\blfootnote{\textsuperscript{*}Core contributors. \quad \textsuperscript{\dag}Project lead.}

\begin{abstract}
How can vision-language-action (VLA) models adapt to new environments where world dynamics shift? While recent research has combined world modeling and action prediction to improve VLA performance, existing methods largely rely on pretraining on static datasets, without mechanisms for active adaptation at deployment time. As a result, these models often fail to generalize when deployed in unseen scenarios with novel object configurations or dynamics.
We present \textbf{WorldAgen}, a unified framework that jointly learns world modeling and action prediction while enabling \textbf{Test-Time Training (TTT)} to adapt to new environments. WorldAgen employs a shared Transformer backbone with two heads: (1) a \textbf{world model head} that predicts future states from past state-action trajectories, and (2) an \textbf{agent model head} that predicts actions conditioned on task instructions. We design a \textbf{Mixed Unidirectional Attention Mask} to separate these two models. During test time, WorldAgen samples exploratory actions, collects ground-truth state transitions, and performs lightweight TTT updates to refine its world model. This adaptation improves the model's understanding of the environment and leads to more accurate action predictions.
Experiments on the CALVIN and LIBERO benchmarks demonstrate that our baseline model achieves comparable, and in some cases superior, performance to current state-of-the-art approaches. Moreover, with TTT on a small number of samples, our method surpasses existing state-of-the-art models, highlighting the effectiveness of adapting world models at inference time.
\end{abstract}

\section{Introduction}

\begin{figure*}[!htbp]
  \centering

\includegraphics[width=\textwidth]{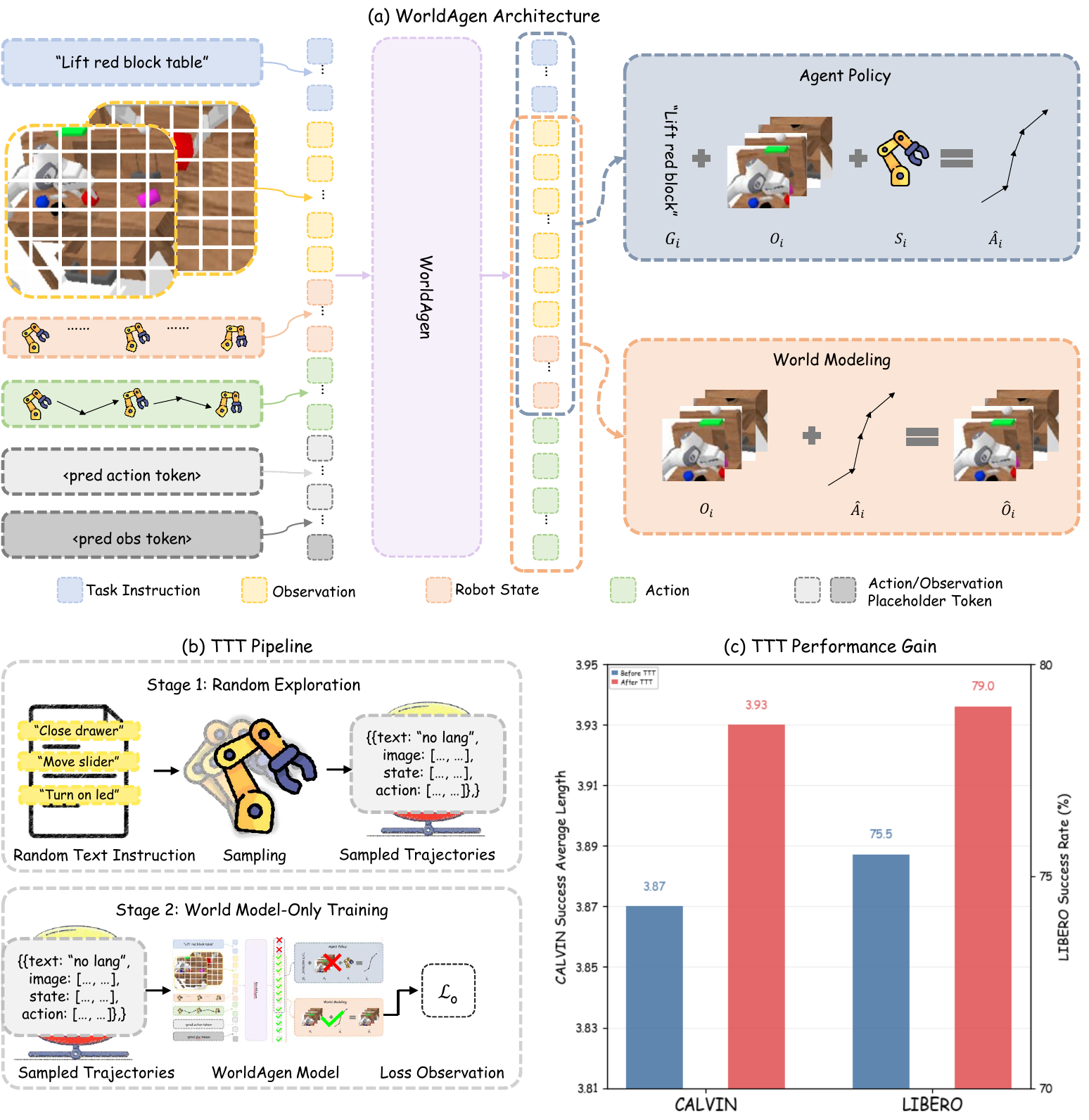}
  
  \caption{
    Overview of WorldAgen.
    (a) WorldAgen Architecture. WorldAgen unifies an agent policy head for task-conditioned action prediction and a world modeling head for task-agnostic state prediction within a single Transformer backbone. At trajectory unit $i$, the agent receives the task instruction $G_i$, the observation chunk $O_i$, and the robot state chunk $S_i$, the action placeholder $A_p$ and observation placeholder $O_p$. Then it predicts the action chunk $\hat{A}_i$. After that, the world model predicts the future observation chunk $\hat{O}_i$ to refine the shared representation of environment dynamics.
    (b) Two-stage TTT pipeline. Stage 1 performs random exploration in the new environment to collect unlabeled trajectories. Stage 2 adapts the world model using LoRA-based fine-tuning on the observation loss $\mathcal{L}_o$, improving environment modeling without modifying the agent policy. 
    (c) TTT Performance. We tested WorldAgen with TTT strategy on CALVIN and LIBERO benchmark, which show great performance gain. 
    }
    \label{fig:worldagen_overview}
\end{figure*}
Vision-language-action (VLA) models have emerged as a powerful and popular paradigm for robotic manipulation \citep{ma2025surveyvisionlanguageactionmodelsembodied, din2025visionlanguageactionmodels}, enabling agents to follow natural language instructions and act directly from raw visual observations.
Recent work has explored joint state-action prediction techniques \citep{tian2024predictiveinversedynamicsmodels,guo2024prediction,liu2025rdt1bdiffusionfoundationmodel,xian2023chaineddiffuser,wolf2025diffusionmodelsroboticmanipulation}, allowing models not only to predict the next actions but also to anticipate how their actions will transform the environment.
This joint formulation has improved data efficiency and scene understanding, leading to stronger performance across manipulation benchmarks.

However, we argue that current methods remain fundamentally limited by training on static datasets \citep{kim2025finetuningvisionlanguageactionmodelsoptimizing,li2025controlvlafewshotobjectcentricadaptation,kim2024openvlaopensourcevisionlanguageactionmodel}.
Once pretrained, these models lack mechanisms to adapt their internal representations of world dynamics when deployed in novel environments.
In realistic settings, distribution shifts such as new object layouts, lighting conditions, or physical properties are inevitable, and static world modeling fails to capture these variations \citep{wu2023daydreamer,lanier2025adaptingworldmodelslatentstate}.
This leads to a key question:
\textbf{How can we enable VLA models to actively adapt their understanding of the environment during test time?}

As shown in Figure \ref{fig:worldagen_overview}, we address this question with WorldAgen, a unified framework that combines joint state-action prediction with a novel TTT strategy:

\begin{itemize}
    \item \textbf{Joint State-Action Modeling.} WorldAgen uses a shared Transformer backbone with two heads. The world model head predicts future states from historical state-action trajectories, while the policy head predicts actions conditioned on task instructions. By training these tasks jointly, WorldAgen aligns scene understanding with action prediction.
    \item \textbf{Test-Time Training for Scene Adaptation.} During test time, WorldAgen executes exploratory actions to collect ground-truth state transitions. It then performs lightweight, LoRA-based TTT updates to refine its world model. These updates improve the internal representation of the environment, indirectly boosting the agent's ability to generate effective actions in novel scenarios, as depicted in Figure~\ref{fig:worldagen_overview}(b).
\end{itemize}
Our work reframes world modeling from a passive pretraining objective into an active test time adaptation mechanism, bridging the gap between offline training and real-world generalization.

In sum, we make the following contributions:
\begin{itemize}
    \item \textbf{Unified joint state-action prediction architecture.} We present a unified Transformer-based structure that integrates world modeling and action prediction, enabling shared representations and work separately via a Mixed Unidirectional Attention Mask.
    \item \textbf{Active test-time adaptation for VLA models.} We introduce a TTT paradigm that transforms world modeling from a static pretraining into an active, test time adaptation mechanism, allowing VLA models to refine their scene understanding on the fly.
   \item \textbf{Empirical validation across challenging benchmarks.} Our baseline achieves performance comparable to, or even better than, state-of-the-art methods. Furthermore, by fine-tuning the world modeling component with only a small number of samples at test time, our method attains state-of-the-art results on both the CALVIN and LIBERO benchmarks.

\end{itemize}
We believe these results demonstrate that continuous adaptation during test time, rather than merely scaling model size or pretraining data, is a key ingredient for robust and generalizable robotic manipulation.
\section{Method}

\subsection{Task Formulation}
\label{sec:task-formulation}

We frame our task as a Partially Observable Markov Decision Process (POMDP). Each rollout trajectory is divided into a sequence of trajectory units indexed by $i$. A trajectory unit is defined as 
$U_i = (G_i, O_i, S_i, A_i, A_p, O_p)$, 
where $G_i$ denotes the task instruction, and $O_i$, $S_i$, and $A_i$ denote the observation, robot state, and action chunks, respectively, accompanied by an action placeholder $A_p$ and an observation placeholder $O_p$.

Each chunk (i.e., $O_i$, $S_i$, and $A_i$) consists of a sequence of time-indexed elements, including observations, states, or actions extracted from the raw trajectory $\tau$. Each element (i.e., $o_t$, $s_t$, $a_t$) is further tokenized into a sequence of tokens that serves as the actual model input.

In unit~$i$, the model first predicts an action chunk $\hat{A}_i$
conditioned on the instruction $G_i$, the observation chunk $O_i$, 
and the state chunk $S_i$.
It then predicts the next observation chunk conditioned on 
$(O_i, S_i, A_i)$ during training or $(O_i, S_i, \hat{A}_i)$ during inference\footnote{A history buffer $h$ is also included in the conditioning; detailed formulation is provided in a later section.}.
Executing the predicted actions advances the environment and produces 
the next unit $U_{i+1}$.
A summary of the key notations used in the WorldAgen framework is 
provided in Table~\ref{tab:notation} in the Appendix.
\subsection{Joint State--Action Prediction}
\label{sec:join-action}

WorldAgen contains two predictive components: a task-conditioned agent model and a task-agnostic world model. Both models share a single Transformer backbone, enabling a unified representation across tasks, as illustrated in Figure~\ref{fig:worldagen_overview}(a).

We maintain two types of histories:

\textbf{Agent history:}
\[
h^a_i = (G_i, O_x, S_x, \hat{A}_x)_{x=i-k}^{i-1},
\]
which contains the task instruction $G_i$ along with the observation, robot state, and predicted action chunks from the previous $k$ segments. This history is used by the agent model for task-aware action prediction.

\textbf{World history:}
\[
h^w_i = (O_x, S_x, \hat{A}_x)_{x=i-k}^{i-1},
\]
which includes the observation, robot state, and predicted action chunks from the previous $k$ segments, and is used by the world model to predict task-agnostic environment dynamics.

We define the two models at trajectory segment $U_i$ as follows:

\textbf{Agent Model.}
The agent model predicts the next action chunk $\hat{A}_i$ conditioned on the current observation chunk $O_i$, robot state chunk $S_i$, task instruction $G_i$, and agent history $h^a_i$:
\[
p_a\!\left(
  \hat{A}_i
  \,\big\vert\,
  G_i, O_i, S_i, h^a_i
\right).
\]

\textbf{World Model.}
The world model predicts the next observation chunk $\hat{O}_i$ conditioned on the current observation chunk $O_i$, robot state chunk $S_i$, predicted action chunk $\hat{A}_i$, and world history $h^w_i$:
\[
p_w\!\left(
  \hat{O}_i
  \,\big\vert\,
  O_i, S_i, \hat{A}_i, h^w_i
\right).
\]

During training, we adopt teacher forcing and replace all predicted actions $\hat{A}_i$ with the ground-truth actions $A_i$.

\begin{figure*}[t]
    \centering
    \includegraphics[scale=0.55]{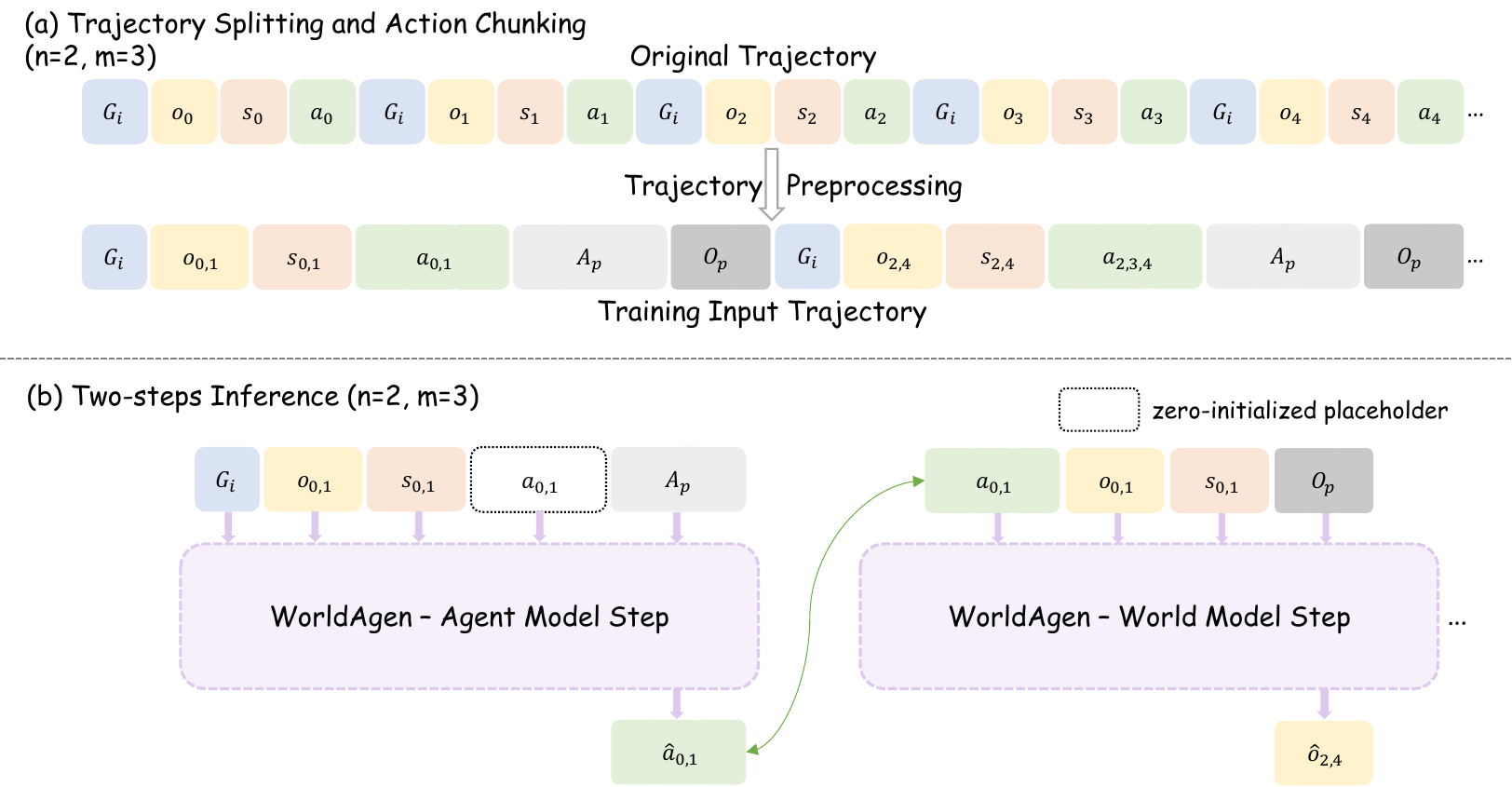}
    \caption{
\textbf{Trajectory Splitting, Chunking, and Two-Step Inference.}
Example with observation chunk length $n=2$ and action chunk length $m=3$.
(a) \textit{Trajectory Splitting and Chunking:}
a raw trajectory is divided into trajectory units $U_i$, where each unit
contains an observation chunk $O_i$ and a robot state chunk $S_i$ obtained
via uniform subsampling from $m$ consecutive observations and states, and an
action chunk $A_i$ obtained by grouping $m$ actions.
(b) \textit{Two-Step Inference:}
the agent model first initializes a zero action placeholder $A_p$ and then
replaces it with the predicted action chunk $\hat{A}_i$.
The world model subsequently predicts the next observation chunk
$\hat{O}_i$, conditioned on the updated trajectory unit.
}
    \label{fig:worldagen}
\end{figure*}

\subsection{Trajectory Splitting and Action Chunking}
\label{traj_split}
Predicting multiple actions at once has been shown to
improve the execution efficiency of real-world robots \cite{intelligence2025pi_}. However, consecutive
observation frames often contain highly redundant visual information, and predicting long observation sequences slows down both training and deployment. To balance these factors, WorldAgen supports different action chunk lengths $m$ and observations chunk length $n$.
We further propose a robust trajectory splitting and action chunking scheme that flexibly adapts to different experimental configurations.

Concretely, given a rollout $\tau = \{(o_t, s_t, a_t)\}_{t=0}^{T-1}$ with $T$ timesteps, we
transform it into a sequence of trajectory units $\{U_i\}_{i=0}^{I-1}$ with $I$ units.
The initial unit $U_0$ aggregates the first $n$ timesteps of the rollout and is paired
with the first action chunk $A_0 = A_{0 \rightarrow n-1} = [a_0, a_1, \ldots, a_{n-1}]$.
For units $U_i$ with $i>0$, we uniformly sample $n$ timesteps to construct the current
observation chunk $O_i$, state chunk $S_i$, and associated action chunk $A_i$:
\begin{align*}
O_i &= O^{n}_{n+(i-1)m \rightarrow n+im-1} \\
    &= \big[
             o_{\,n+(i-1)m},\,
             o_{\,n+(i-1)m + \lfloor \tfrac{1}{n} m \rfloor},\,
             \ldots,\,
             o_{\,n+(i-1)m + \lfloor \tfrac{n-1}{n} m \rfloor}
       \big], \\
S_i &= S^{n}_{n+(i-1)m \rightarrow n+im-1} \\
    &= \big[
          s_{\,n+(i-1)m},\,
          s_{\,n+(i-1)m + \big\lfloor \tfrac{1}{n} m \big\rfloor},\,
\\
    &\quad
          \ldots,\,
          s_{\,n+(i-1)m + \big\lfloor \tfrac{n-1}{n} m \big\rfloor}
       \big], \\
A_i &= A_{n+(i-1)m \rightarrow n+im-1} \\
    &= \big[
             a_{\,n+(i-1)m},\,
             a_{\,n+(i-1)m+1},\,
             \ldots,\,
             a_{\,n+im-1}
       \big].
\end{align*}
This procedure keeps each observation chunk fixed to length $n$, reducing redundant
visual information and improving runtime efficiency.

Figure~\ref{fig:worldagen} (a) illustrates an example with $n=2$ and $m=3$. Algorithm of trajectory splitting and action chunking for variable length of action, observation and robot state chunk is shown in Algorithm~\ref{alg:traj-splitting} in the Appendix, where $\textsc{UniformSubsample}(a,b,n)$ uniformly samples $n$ distinct indices from the integer interval $\{a,\ldots,b\}$ without replacement and returns them in ascending order, which we then use to construct the
observation and state chunks.

\begin{figure}[t]
    \centering
    \includegraphics[scale=0.5]{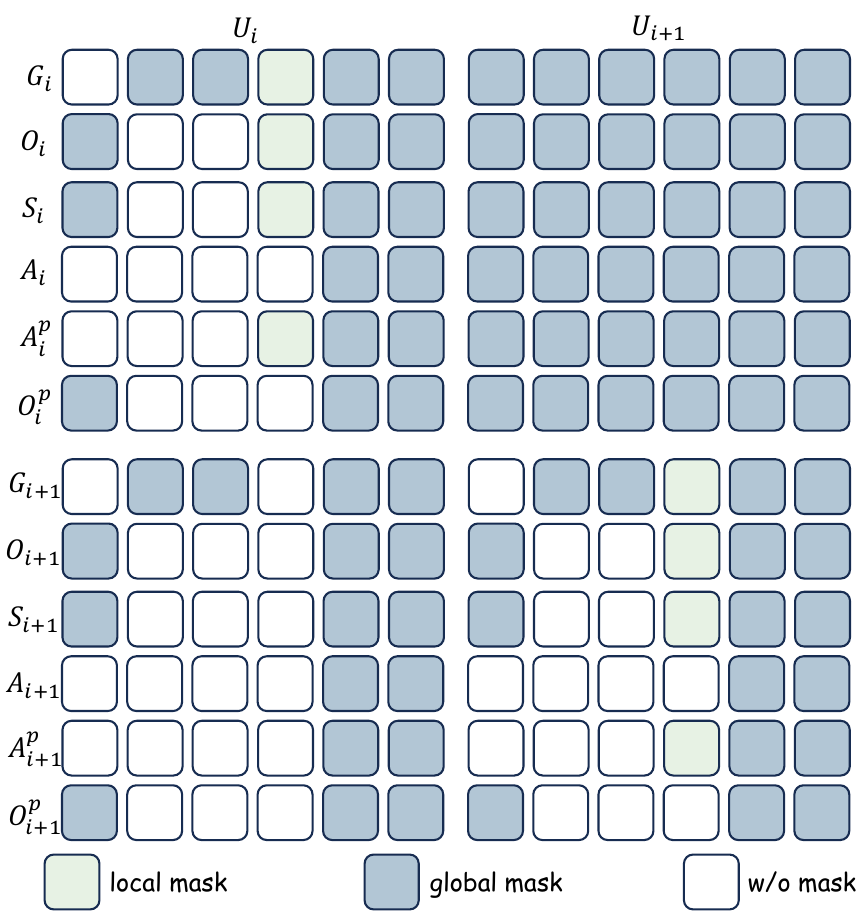}
    \caption{
Proposed Mixed Unidirectional Attention Mask: combines local and global masking. For each trajectory unit, WorldAgen applies a local mask that prevents the action chunk $A_p$ from attending to the corresponding action placeholder $A_i$. A global mask is further used to construct a task-agnostic world model head, so that the observation placeholder $O_p$ will not attend to the task instruction $G_i$. General structure conforms to strictly causal prediction.
}
    \label{fig:attn_mask}
\end{figure}

\subsection{Mixed Unidirectional Attention Mask}
To ensure that the world modeling and agent policy tasks are trained jointly without information leakage, we introduce a mixed unidirectional attention masking mechanism as shown in Figure \ref{fig:attn_mask}. 
This mechanism integrates two complementary components:

\textbf{Local mask:} ensures that the current observation chunk $O_i$, the robot proprioception chunk $S_i$, the task instruction $G_i$, and the action placeholder $A_p$ cannot access the action chunk $A_i$. This prevents information leakage within a single timestep.

\textbf{Global mask:} guarantees that the world model head remains invisible to the task instruction $G_i$ across all time steps. Additionally, the global mask enforces mutual invisibility among the task instruction and observations to avoid cross-modal leakage. 

Meanwhile, the general structure conforms to strictly causal prediction, such that both the world model and policy heads cannot access any future tokens.

\subsection{Pretraining}
We pretrain WorldAgen using large-scale trajectory data collected from robot demonstrations. We split trajectories using our proposed robust splitting method.
During pretraining, we jointly optimize the agent model and world model using teacher forcing:

\[
\mathcal{L} = \mathcal{L}_a + \lambda \mathcal{L}_o,
\]
where $\mathcal{L}_a$ is the cross-entropy loss for predicting future action chunks, $\mathcal{L}_o$ is the reconstruction loss for predicting future observation chunks, and $\lambda$ is a weighting factor balancing the two objectives.

It is worth noting that during training, we feed the ground-truth action chunk (teacher forcing), whereas during testing, the model performs autoregressive prediction by feeding the previously predicted action chunk at each timestep.

\subsection{Two-Step Inference}

At test time, WorldAgen runs in a two-step inference loop over trajectory units as shown in Figure \ref{fig:worldagen} (b). For unit $i$, we first perform the agent model step. We construct
the input sequence with a zero-initialized action placeholder $A_p$ and feed it to the agent model, which predicts the current action chunk $\hat{A}_i$. The predicted actions then overwrite the placeholder $A_p$.

Next, we perform the world model step. Conditioned on the task instruction $G_i$, the current observation chunk
$O_i$ and the updated action chunk $A_i$ (same as $\hat{A}_i$), the world model predicts the next observation chunk $\hat{O}_i$. During testing, these two steps are applied alternately to roll out the whole trajectory.

\subsection{Test-Time Training}
While joint pretraining improves world understanding, distribution shifts in novel environments can still degrade performance.
To address this, we introduce a two-stage Test-Time Training strategy that adapts the world model online, refining its understanding of environment dynamics and indirectly improving downstream action prediction. The whole pipeline is given in Figure~\ref{fig:worldagen_overview} (b).

\paragraph{Stage 1: Random Exploration and Data Collection.}
At deployment, the agent first performs free exploratory rollouts in the target environment.
During this phase, it records trajectories of $(O_i, S_i, A_i)$.
To ensure that adaptation focuses on environment dynamics rather than task-specific instructions, all collected trajectories are relabeled with a generic ``no-lang'' token in place of language instructions.

\paragraph{Stage 2: World Model Adaptation.}
Using the collected trajectories, we adapt only the shared backbone with LoRA-based parameter-efficient fine-tuning, while keeping the policy head frozen.  
The update rule is:
\[
\theta^\prime_w \leftarrow \theta_w - \eta \nabla_{\theta_w} \mathcal{L}_o,
\]
where $\theta_w$ are the parameters of the world model head and $\eta$ is the learning rate.

This targeted update improves world understanding ability and enhances the shared representations, which are in turn leveraged by the agent model for action prediction.

By decoupling adaptation from the task goal and restricting it to the world model, \textbf{our TTT procedure enables robust scene adaptation without requiring additional task-specific annotations, paving the way for scalable test-time adaptation in VLA models.}

\begin{table}[!t]
\centering
\small
\setlength{\tabcolsep}{4pt}
\begin{tabular}{l|ccccc|c}
\toprule
Method & T1 & T2 & T3 & T4 & T5 & Avg. Len. $\uparrow$ \\
\midrule
RoboFlamingo      & 82.4 & 61.9 & 46.6 & 33.1 & 23.5 & 2.47 \\
SuSIE             & 87.0 & 69.0 & 49.0 & 38.0 & 26.0 & 2.69 \\
GR-1              & 85.4 & 71.2 & 59.6 & 49.7 & 40.1 & 3.06 \\
3D Diffuser Actor & 92.2 & 78.7 & 63.9 & 51.2 & 41.2 & 3.27 \\
CLOVER            & 96.0 & 83.5 & 70.8 & 57.5 & 45.4 & 3.53 \\
Seer              & 93.0 & 82.4 & 72.3 & 62.6 & 53.3 & 3.64 \\
Seer-Large        & 92.7 & 84.6 & 76.1 & 68.9 & 60.3 & 3.83 \\
\midrule
\textbf{WorldAgen}        & 96.3 & 87.7 & 76.8 & 67.3 & 59.1 & 3.87 \\
\textbf{WorldAgen-TTT}    & \textbf{96.6} & \textbf{88.5} & \textbf{78.5} & \textbf{68.7} & \textbf{60.5} & \textbf{3.93} \\
\bottomrule
\end{tabular}
\caption{Performance comparison on the CALVIN benchmark.
We report the success rate (\%) for completing 5 consecutive tasks 
and the average sequence length (ASL).}
\label{tab:calvin_results}
\end{table}

\section{Experiments}

\subsection{Datasets}
\paragraph{CALVIN.} CALVIN \citep{mees2022calvin} is an open-source simulated benchmark designed for learning long-horizon language-conditioned robot manipulation tasks. The benchmark requires agents to solve complex manipulation tasks by understanding a series of unconstrained language instructions in sequence.

\paragraph{LIBERO.} LIBERO \citep{liu2023libero} is a comprehensive benchmark for lifelong learning in robot manipulation that emphasizes knowledge transfer across diverse tasks.

\subsection{Implementation Details}

\paragraph{Test-Time Training} TTT represents a crucial component of WorldAgen that enables adaptive performance improvements during inference. In the TTT phase, we apply LoRA fine-tuning to the backbone of Qwen3. Specifically, we apply LoRA to the attention projection layers (q\_proj, k\_proj, v\_proj, o\_proj) and the MLP projection layers (gate\_proj, up\_proj, down\_proj) of Qwen3.

The test-time training procedure, including sampling and LoRA adaptation, can be executed on an RTX 4090 GPU and requires approximately 8 minutes per task for CALVIN and 2 minutes per scene for LIBERO. More details can be found in Appendix.

\subsection{Results}
\label{sec:main_results}

\begin{table*}[t]
\centering

\begin{tabular}{>{\centering\arraybackslash}m{3.2cm}>{\centering\arraybackslash}m{2.0cm}>{\centering\arraybackslash}m{2.0cm}>{\centering\arraybackslash}m{2.0cm}>{\centering\arraybackslash}m{2.0cm}>{\centering\arraybackslash}m{2.2cm}}
\toprule
Method & 
\begin{tabular}{c} Avg. \\ Success $\uparrow$ \end{tabular} & 
\begin{tabular}{c} Put soup \\ and box \\ in basket \end{tabular} & 
\begin{tabular}{c} Put box \\ and butter \\ in basket \end{tabular} & 
\begin{tabular}{c} Turn on \\ stove and \\ put pot \end{tabular} & 
\begin{tabular}{c} Put bowl in \\ drawer and \\ close it \end{tabular} \\
\midrule
MT-ACT & 41.0 & 30.0 & 50.0 & 75.0 & 85.0 \\
MVP & 68.2 & 83.3 & \textbf{90.0} & 80.0 & 88.3 \\
MPI & 77.3 & 66.6 & 86.6 & 96.6 & 95.0 \\
OpenVLA & 54.0 & 35.0 & 95.0 & 65.0 & 45.0 \\
Seer & 78.7 & 80.0 & \textbf{90.0} & 91.7 & 81.7 \\
\midrule
WorldAgen & 75.5 & 70.0 & 75.0 & 95.0 & 100 \\
\textbf{WorldAgen-TTT} & \textbf{79.0} & \textbf{85.0} & 75.0 & \textbf{95.0} & \textbf{100} \\
\bottomrule
\end{tabular}

\vspace{0.2cm}

\begin{tabular}{>{\centering\arraybackslash}m{2.5cm}>{\centering\arraybackslash}m{2.2cm}>{\centering\arraybackslash}m{2.2cm}>{\centering\arraybackslash}m{2.2cm}>{\centering\arraybackslash}m{2.2cm}>{\centering\arraybackslash}m{2.4cm}}
\toprule
\begin{tabular}{c} Put mugs on \\ left and \\ right plates \end{tabular} & 
\begin{tabular}{c} Pick book \\ and place it \\ in back \end{tabular} & 
\begin{tabular}{c} Put mug on \\ plate and put \\ pudding to right \end{tabular} & 
\begin{tabular}{c} Put soup \\ and sauce \\ in basket \end{tabular} & 
\begin{tabular}{c} Put both pots \\ on stove \end{tabular} & 
\begin{tabular}{c} Put mug in \\ microwave and \\ close it \end{tabular} \\
\midrule
20.0 & 75.0 & 0.0 & 0.0 & 10.0 & 65.0 \\
46.7 & 63.3 & 45.0 & 78.3 & \textbf{60.0} & 46.7 \\
83.3 & 83.3 & 56.6 & 86.6 & 40.0 & 78.3 \\
40.0 & 80.0 & 60.0 & 45.0 & 20.0 & 55.0 \\
85.0 & 65.0 & \textbf{86.7} & 88.3 & 51.7 & \textbf{66.7} \\
\midrule
85.0 & 90.0 & 60.0 & 100 & 45.0 & 35.0 \\
\textbf{90.0} & \textbf{90.0} & 50.0 & \textbf{100} & 45.0 & \textbf{60.0} \\
\bottomrule
\end{tabular}
\caption{Performance comparison on LIBERO benchmark. We report the success rate (\%) across different manipulation tasks. }
\label{tab:libero_results}
\end{table*}

Our experimental results demonstrate the effectiveness of WorldAgen across both CALVIN and LIBERO benchmarks. 

\noindent \textbf{CALVIN} As shown in Table~\ref{tab:calvin_results}, we compare WorldAgen with recent VLA baselines, including RoboFlamingo~\cite{li2023vision}, SuSIE~\cite{black2023zero}, GR-1~\cite{wu2023unleashing}, 3D Diffuser Actor~\cite{ke20243d}, and CLOVER~\cite{bu2024closed}.  WorldAgen follows a GR-1-style GPT video architecture, but introduces the Mixed Unidirectional Attention Mask to separate task instruction from world modeling and employs a lightweight TTT strategy to enhance world-modeling capability. Our method achieves consistent performance improvements across five consecutive tasks, indicating that \textbf{\textit{strengthening world modeling via TTT improves environment understanding and boosts long-horizon manipulation performance.}}

\noindent \textbf{LIBERO} As shown in Table~\ref{tab:libero_results}, we further evaluate WorldAgen on the LIBERO benchmark against multi-task VLA baselines such as MT-ACT~\cite{bharadhwaj2024roboagent}, MVP~\cite{xiao2022masked}, MPI~\cite{zeng2024learning}, OpenVLA~\cite{kim2024openvlaopensourcevisionlanguageactionmodel} and Seer~\cite{tian2024predictiveinversedynamicsmodels}. WorldAgen attains the best or on-par performance across most LIBERO subsets, further supporting that test-time adaptation of the world model leads to stronger environment understanding and improved robustness to distribution shifts.

We observe improvements or maintained performance across almost every task compared to the baseline. These consistent gains across diverse manipulation scenarios mirror our findings on CALVIN, demonstrating \textbf{\textit{robustness and generalization ability}} of our TTT approach across different benchmarks and scenario dynamics.

\subsection{Ablation Study}
\label{sec:ablation}

To better understand the key components contributing to performance, we conduct ablation studies on (1) the contribution of world modeling to policy learning, (2) the effect of LoRA parameterization during TTT, (3) the impact of the amount of randomly sampled data used for TTT, and (4) transfer from simulator to the real robot. We further analyze the impact of trajectory splitting and preprocessing, image and action chunk lengths, and different Transformer backbones, and compare LoRA-based TTT with full fine-tuning while visualizing performance before and after TTT. All ablation and additional experimental results are provided in the Appendix.

\subsubsection{World Modeling Ability}

\begin{table}[h]
\centering
\begin{tabular}{lcc}
\toprule
Dataset & World Modeling & Avg. Success $\uparrow$ \\
\midrule
\multirow{2}{*}{CALVIN} & \text{\texttimes} & 2.96 \\ & \checkmark & \textbf{3.87} \\
\midrule
\multirow{2}{*}{LIBERO} & \text{\texttimes} & 46.5\% \\ & \checkmark & \textbf{75.5\%} \\
\bottomrule
\end{tabular}
\caption{Ablation study on world modeling. We compare models with and without image prediction to evaluate the contribution of world modeling to agent policy learning.}
\label{tab:world_model_ablation}
\end{table}

To validate the contribution of world modeling to policy learning, we compare models with and without image prediction capability by including or removing image-prediction tokens in the input as shown in Table~\ref{tab:world_model_ablation}. World modeling consistently improves performance: on CALVIN, the average success length increases from 2.96 to 3.87 (+30.7\%), and on LIBERO, the success rate rises from 46.5\% to 78.0\% (+67.7\%). These results indicate that \textbf{\textit{world modeling yields richer representations of environment dynamics, leading to more effective action prediction and policy learning.}}

\subsubsection{LoRA Configuration}
\begin{table}[h]
\centering
\begin{tabular}{cc}
\toprule
LoRA Rank & Avg. Len. $\uparrow$ \\
\midrule
16 & 3.928 \\
32 & 3.918 \\
64 & 3.918 \\
\textbf{128} & \textbf{3.930} \\
256 & 3.923 \\
\bottomrule
\end{tabular}
\caption{Ablation study on LoRA rank during TTT in CALVIN dataset. All other parameters are kept constant.}
\label{tab:lora_ablation}
\end{table}

We study the effect of LoRA rank on TTT performance while fixing the learning rate and training data size as shown in Table~\ref{tab:lora_ablation}. Performance only varies within 0.02 across ranks, indicating that \textbf{\textit{under fixed hyperparameters, TTT performance is largely insensitive to the LoRA rank.}} We therefore use a LoRA rank of 128 in our main experiments, which slightly outperforms other settings while remaining computationally efficient.

\subsubsection{TTT Data Sampling Volume}
\begin{table}[h]
\centering
\begin{tabular}{cc}
\toprule
Test Time Training Data & Avg. Lens. $\uparrow$ \\
\midrule
6 & 3.871 \\
90 & 3.922 \\
\textbf{204} & \textbf{3.928} \\
340 & 3.917 \\
\bottomrule
\end{tabular}
\caption{Ablation study on TTT data sampling volume. We vary the amount of test-time training data, defined as the product of number of samples and repeat times.}
\label{tab:data_volume_ablation}
\end{table}

We study how the amount of test-time training data affects performance as shown in Table~\ref{tab:data_volume_ablation}. As the number of TTT samples increases from 6 to 204, the average success score improves from 3.871 to 3.928, indicating that more diverse exploration data benefits world-model adaptation. However, further increasing the data to 340 slightly reduces performance to 3.917, suggesting diminishing returns and mild overfitting to image generation rather than action prediction. These results indicate that \textbf{\textit{moderately increasing TTT data improves performance, but excessive data can be counterproductive, highlighting the need for an appropriate TTT data budget.}}

\subsubsection{Simulator to Real World}

\begin{table}[h]
\centering

\begin{tabular}{lccc}
\toprule
\textbf{} & w/o TTT & w/o Noise & w/ Noise \\
\midrule
CALVIN & 3.87 & 3.93 & 3.90 \\
LIBERO & 75.5 & 79.0 & 78.0 \\
\bottomrule
\end{tabular} 
\caption{Performance comparison under Gaussian noise perturbation.}
\label{tab:gaussian_noise}
\end{table}

We add Gaussian noise (std = 0.1) to the observations to simulate real-world conditions, such as camera distortion and sensor noise, as shown in Table~\ref{tab:gaussian_noise}. 

This experiment indicates that \textbf{\textit{even under noisy conditions, our method consistently outperforms the baselines, demonstrating its robustness and strong adaptability to real-world uncertainty.}}
\section{Related Work}
\paragraph{Vision-Language-Action Models.}
Vision-Language-Action (VLA) models unify perception, language, and control for robotic manipulation \citep{sapkota2025visionlanguageactionmodelsconceptsprogress,din2025visionlanguageactionmodels}. 
Early systems such as RT-1 and RT-2 established the effectiveness of large-scale transformer policies, while PaLM-E and LLaVA further incorporated multimodal grounding \citep{brohan2023rt1roboticstransformerrealworld,zitkovich2023rt,driess2023palm,liu2023visual}. 
However, most VLA models directly predict actions without explicitly modeling world dynamics, limiting generalization to novel environments \citep{zhang2025inspirevisionlanguageactionmodelsintrinsic}.

\paragraph{World Models.}
World models learn predictive dynamics to support planning and decision-making \citep{sutton1990integrated,hafner2025mastering}. 
Neural approaches such as World Models and Dreamer improve sample efficiency and long-horizon reasoning \citep{ha2018world,hafner2022masteringataridiscreteworld,hafner2025mastering}. 
In robotics, however, these models are typically used for state prediction or model-based planning and remain largely decoupled from language-grounded action generation \citep{sakagami2023robotic, wang2026vagen}.

\paragraph{Test-Time Training.}
Test-time training adapts models to distribution shifts during inference via self-supervised objectives \citep{sun2020testtimetrainingselfsupervisiongeneralization,ma2024improved}. 
While TTT has been explored in both vision and NLP \citep{ye2023robustquestionansweringdistribution}, robotic adaptation has mainly focused on perception modules or low-level policies \citep{zhao2020sim}. 
Our work instead incorporates TTT into a unified VLA framework by adapting the world model itself during deployment.

\section{Conclusion}

We introduced \textbf{WorldAgen}, a unified vision–language–action framework that jointly learns world modeling and action prediction through a shared Transformer backbone. WorldAgen integrates a world model head that predicts future states and a policy head that predicts task-conditioned actions. At test time, the model performs lightweight Test-Time Training (TTT) using short exploratory rollouts, turning world modeling into an online adaptation signal. This enables the agent to quickly adjust to new environments and leads to consistent performance gains on CALVIN and LIBERO, demonstrating that adaptive VLA systems can benefit substantially from continual test-time interaction.

\section{Limitations and Future Work}
Our study evaluates only a single model size and architecture, and focuses test-time training exclusively on the world-model component. Future work includes exploring larger and more diverse model architectures, as well as developing joint test-time-training methods that improve both action prediction and world modeling.

\bibliography{aaai2026}

\appendix
\newpage
\renewcommand{\partname}{}
\part{Appendix}

This appendix provides comprehensive supplementary materials for our WorldAgen framework. We present detailed implementation specifications including benchmark descriptions, model architecture components, and trajectory processing methods for variable-length chunks. Additionally, we report extensive experimental results examining the effects of different chunk configurations on baseline performance, and comprehensive TTT analyses comparing LoRA \citep{hu2022lora} versus full fine-tuning approaches across various parameter settings and data sizes.

\section{Implementation Details}
\subsection{Benchmark}

\textbf{LIBERO Benchmark} ~\citep{liu2023libero} is a comprehensive benchmark for lifelong learning in robot manipulation that consists of four distinct task suites designed to evaluate different aspects of knowledge transfer. The benchmark includes LIBERO-Spatial (10 tasks focusing on spatial relationship transfer), LIBERO-Object (10 tasks emphasizing object-centric knowledge transfer), LIBERO-Goal (10 tasks targeting procedural knowledge generalization), and LIBERO-100 (100 tasks with highly entangled knowledge requirements). Each task suite is accompanied by high-quality human teleoperation demonstrations to support sample-efficient learning.

We use LIBERO-100, which is the most challenging and comprehensive subset, containing 100 diverse manipulation tasks that require the transfer of mixed declarative and procedural knowledge. This suite is further divided into LIBERO-90, consisting of 90 short-horizon tasks used for pretraining policies, and LIBERO-10 (also referred to as LIBERO-Long), which contains 10 long-horizon tasks specifically selected for evaluating downstream lifelong learning performance.

\textbf{CALVIN Benchmark} ~\citep{mees2022calvin} is an open-source simulated benchmark designed for learning long-horizon language-conditioned robot manipulation tasks. The dataset contains 34 manipulation tasks that are more complex than existing vision-and-language datasets in terms of sequence length, action space, and language complexity. The environment features a Franka Emika Panda robot with a parallel-jaw gripper operating in a desktop workspace containing various interactive objects, including a sliding door, drawer, colored blocks, LED, and light bulb that can be manipulated according to unconstrained natural language instructions.

CALVIN provides four different environments (A, B, C, D) that vary in desk colors and object configurations, enabling evaluation across different visual contexts and supporting flexible specification of sensor suites. The benchmark's evaluation protocol requires agents to solve sequences of up to five consecutive tasks by understanding and executing a series of language instructions, such as \textit{open the drawer, pick up the blue block, rotate the block, push the block into the drawer, open the sliding door.} This sequential task completion paradigm makes CALVIN particularly challenging for assessing the generalization capabilities and long-horizon reasoning of language-conditioned manipulation policies.

\subsection{Model Architecture}

Leveraging the flexible architecture of Transformers, we achieve unified processing of image, language, action, and robot state modalities, supporting variable-length image chunks and action chunks for both input and output.

\textbf{Encoder} In the encoder component, we employ a MAE-pretrained ViT-B \citep{he2022masked} as the visual encoder, utilizing dual-view RGB inputs from static and wrist cameras. Similar to Seer \citep{tian2024predictiveinversedynamicsmodels}, we incorporate a Perceiver Resampler \citep{jaegle2021perceiver}  to reduce the number of image tokens, thereby decreasing computational load and improving efficiency. The Perceiver Resampler is an attention-based feature compression module that uses a set of learnable query tokens to extract the most important information from a large number of image features. Through this approach, it compresses the originally large number of image tokens into a fixed number of compact representations, preserving essential visual information while significantly reducing the computational complexity of subsequent network layers.

For language processing, we utilize the text encoder from CLIP ViT-B/32. For robot state and action, we employ MLPs to project them into the same embedding space. Both robot state and action are represented as 7-dimensional vectors, where the first 6 dimensions encode the arm state representing the end effector's position and orientation, and the last dimension represents the gripper state indicating the open/close status. We use separate MLP layers to process arm state and gripper state independently, then concatenate the results.

We initialize predicted tokens as zero-filled tensors with the same dimensions as the target actions and images to be predicted, and insert them after the input data at each time step.

\textbf{Decoder} In the decoder, we index the predicted tokens based on the input and slice out the image and action tokens. For image tokens, we employ the Vision Transformer encoder architecture \citep{he2022masked} for decoding, followed by a linear layer to predict pixel patches. For the action decoder, similarly, we use an MLP to reduce the action-corresponding vectors to 7 dimensions. For the arm state and gripper state components, we employ different linear layers for decoding. For the gripper state, we apply binary thresholding at 0.5 to represent the gripper's open/close status (0 for closed, 1 for open).

\textbf{Backbone} Qwen3 \citep{yang2025qwen3technicalreport} represents the latest advancement in the Qwen large language model family, comprising both dense and Mixture-of-Experts (MoE) architectures with parameter scales ranging from 0.6 to 235 billion. Utilizing Qwen3 as a backbone architecture offers significant advantages for large language model applications, primarily through its innovative parameter efficiency where Qwen3 dense base models achieve performance comparable to much larger Qwen2.5 models while using  fewer parameters, with Qwen3-4B matching the performance of Qwen2.5-72B-Instruct. The MoE variants provide exceptional computational efficiency, as Qwen3-MoE base models deliver similar performance to Qwen2.5 dense base models while utilizing only 10\% of the active parameters, resulting in significant savings in both training and inference costs.

\begin{figure}[t]
    \centering
    \includegraphics[width=0.48\textwidth]{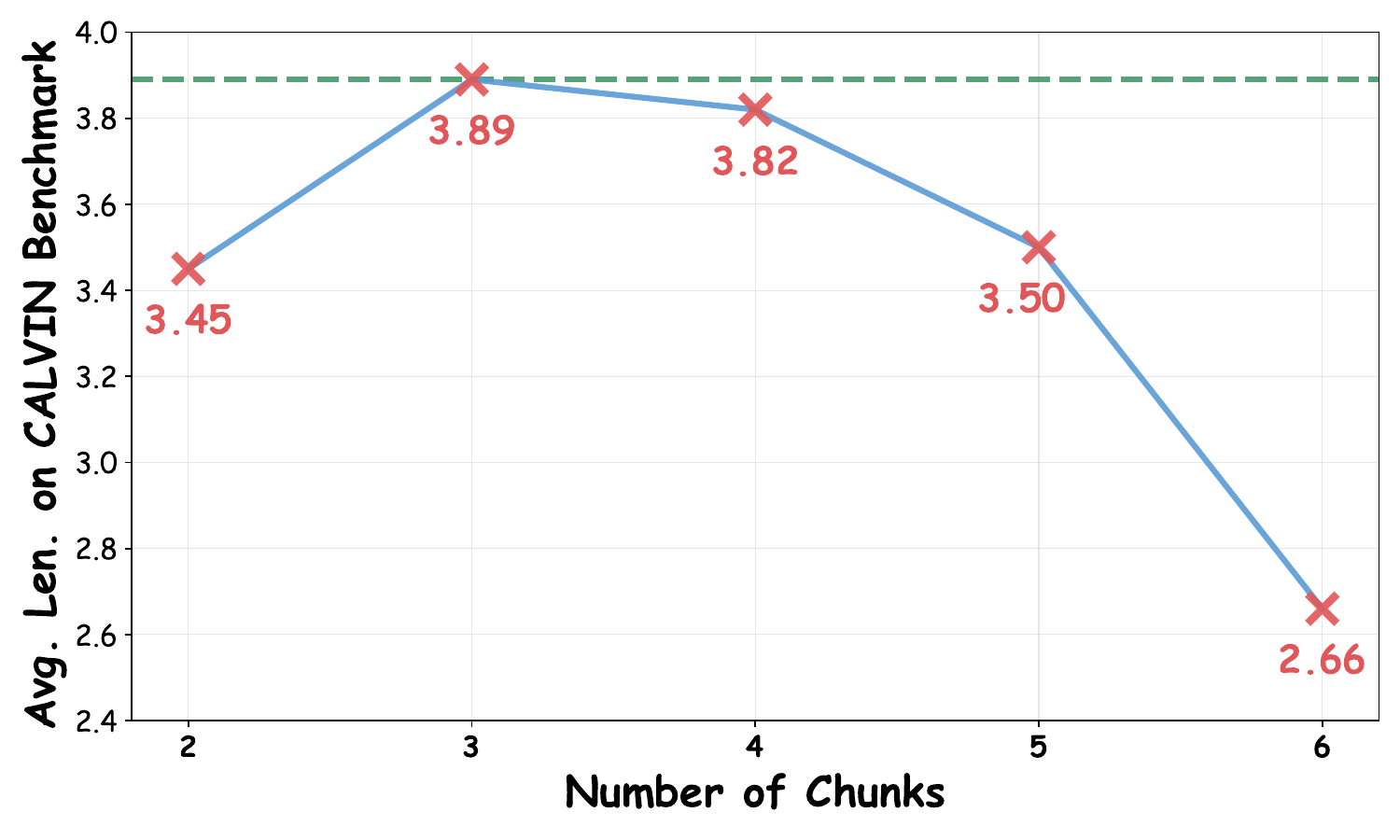}
    \caption{This figure shows the performance of the WorldAgen model on CALVIN as the number of chunks changes.}
    \label{fig:number_of_chunks}
\end{figure}

\section{Baseline Results}

In this section, we present more results of our baseline on CALVIN benchmark, including pretraining, different length and number of chunks and different backbone.

\subsection{Pretraining}
Given the powerful modeling capability and flexibility of Qwen3~\cite{yang2025qwen3technicalreport}, we adopt it as the network backbone for WorldAgen. Following the configuration established in~\citep{tian2024predictiveinversedynamicsmodels}, we configure the transformer architecture with 12 transformer heads and 24 transformer layers, resulting in a total network size of 370M parameters with 120M trainable parameters. 
 Model pretraining is performed on a 4$\times$H100 GPU server, taking approximately 60 hours for CALVIN and 5 hours for LIBERO. 

\begin{figure*}[t]
    \centering
    \includegraphics[width=0.7\textwidth]{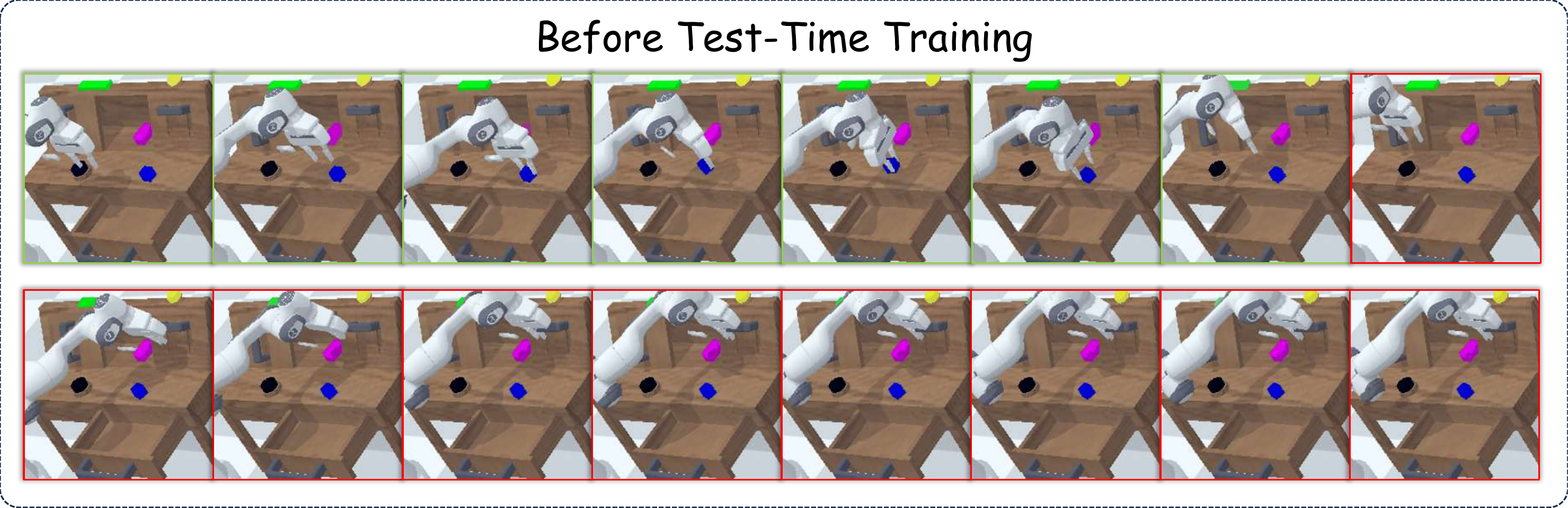}
    \includegraphics[width=0.7\textwidth]{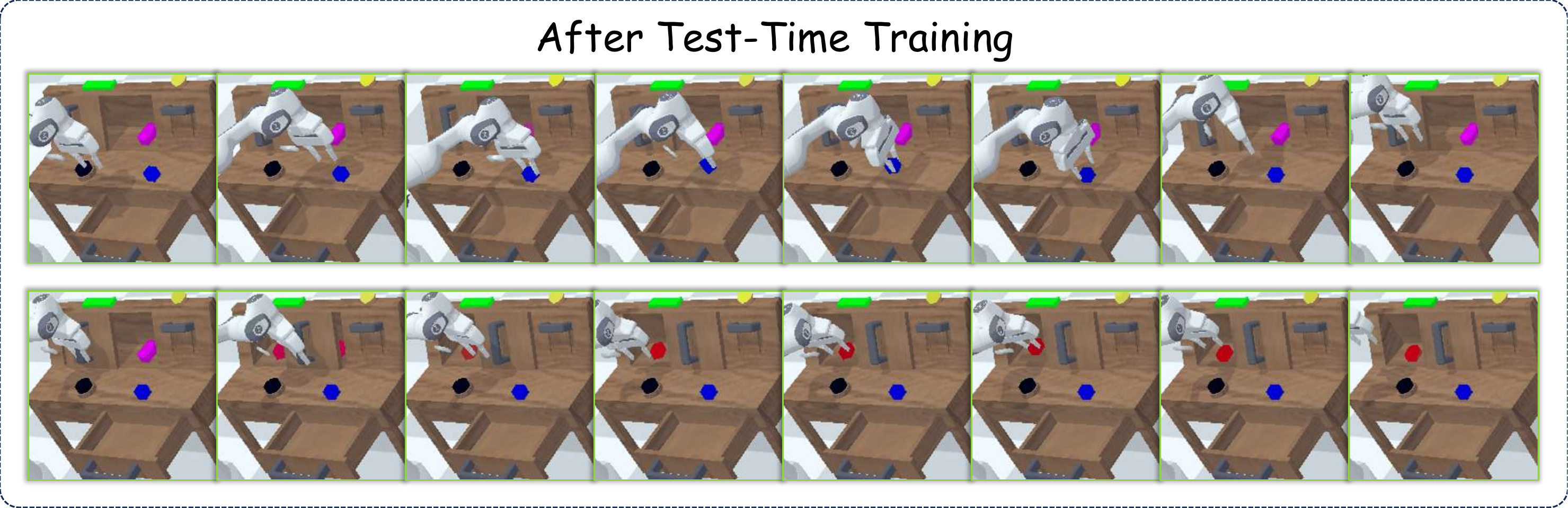}
    \caption{
        In the comparison between the trajectories before and after TTT, the red regions indicate failures within the trajectory, while the green regions represent successful portions.
    }
    \label{fig:demos}
\end{figure*}

\subsection{Image and Action Chunk Lengths}
WorldAgen offers flexible input-output interfaces, accommodating image and action chunks of variable lengths. We evaluate the effect of different image and action chunk lengths on model performance. The results are shown in Table~\ref{tab:chunk_ablation}.

Therefore, on CALVIN benchmark, we employ an image chunk length of 1, action chunk length of 5, and trajectory length of 16. For LIBERO, we use an image chunk length of 1, action chunk length of 3, and trajectory length of 7.

\begin{algorithm}[t]
\caption{Trajectory splitting with variable-length action, observation, and robot state units.}
\label{alg:traj-splitting}
\begin{algorithmic}[1]
\REQUIRE Rollout $\tau = \{(o_t, s_t, a_t)\}_{t=0}^{T-1}$, observation chunk length $n$, action chunk length $m$
\STATE $\mathcal{U} \leftarrow \emptyset$
\FOR{$i = 0$ \TO $I-1$}
  \IF{$i = 0$}
    \STATE $\mathcal{I}^{\text{obs}} \leftarrow \{0,1,\ldots,n-1\}$
    \STATE $O^{n}_{0 \rightarrow n-1} \leftarrow \{\,o_t \mid t \in \mathcal{I}^{\text{obs}}\,\}$
    \STATE $S^{n}_{0 \rightarrow n-1} \leftarrow \{\,s_t \mid t \in \mathcal{I}^{\text{obs}}\,\}$
    \STATE $A_{0 \rightarrow n-1} \leftarrow \{\,a_t \mid t \in \mathcal{I}^{\text{obs}}\,\}$
    \STATE $O_i \leftarrow O^{n}_{0 \rightarrow n-1}$
    \STATE $S_i \leftarrow S^{n}_{0 \rightarrow n-1}$
    \STATE $A_i \leftarrow A_{0 \rightarrow n-1}$
    \STATE $A^p_i \leftarrow \mathbf{0}^{n}$ \hfill // action placeholder (length $n$)
    \STATE $O^p_i \leftarrow \mathbf{0}^{n}$ \hfill // observation placeholder (length $n$)
  \ELSE
    \STATE $a \leftarrow n + (i-1)m$
    \STATE $b \leftarrow n + im - 1$
    \STATE $\mathcal{I}^{\text{obs}} \leftarrow \textsc{UniformSubsample}(a,\, b,\, n)$
    \STATE $O^{n}_{n+(i-1)m \rightarrow n+im-1} \leftarrow \{\,o_t \mid t \in \mathcal{I}^{\text{obs}}\,\}$
    \STATE $S^{n}_{n+(i-1)m \rightarrow n+im-1} \leftarrow \{\,s_t \mid t \in \mathcal{I}^{\text{obs}}\,\}$
    \STATE $A_{n+(i-1)m \rightarrow n+im-1} \leftarrow \{\,a_t \mid n+(i-1)m \le t \le n+im-1\,\}$
    \STATE $O_i \leftarrow O^{n}_{n+(i-1)m \rightarrow n+im-1}$
    \STATE $S_i \leftarrow S^{n}_{n+(i-1)m \rightarrow n+im-1}$
    \STATE $A_i \leftarrow A_{n+(i-1)m \rightarrow n+im-1}$
    \STATE $A^p_i \leftarrow \mathbf{0}^{m}$ \hfill // action placeholder (length $m$)
    \STATE $O^p_i \leftarrow \mathbf{0}^{n}$ \hfill // observation placeholder (length $n$)
  \ENDIF
  \STATE $U_i \leftarrow \bigl(G, O_i, S_i, A_i, A^p_i, O^p_i\bigr)$
  \STATE $\mathcal{U} \leftarrow \mathcal{U} \cup \{U_i\}$
\ENDFOR
\STATE \textbf{return} $\mathcal{U}$
\end{algorithmic}
\end{algorithm}

\begin{table}[h]
\centering
\begin{tabular}{lccc}
\toprule
Dataset & I. Chunk & A. Chunk & Avg. Success $\uparrow$ \\
\midrule
\multirow{6}{*}{CALVIN} & 1 & 3 & 3.82 \\
& \textbf{1} & \textbf{5} & \textbf{3.87} \\
& 1 & 7 & 3.43 \\
& 1 & 9 & 3.16 \\
& 3 & 5 & 3.30 \\
& 5 & 5 & 1.78 \\
\midrule
\multirow{3}{*}{LIBERO} & \textbf{1} & \textbf{3} & \textbf{78.0\%} \\
& 1 & 5 & 74.5\% \\
& 1 & 7 & 47.0\% \\
\bottomrule
\end{tabular}
\caption{Ablation study on image and action chunk lengths across CALVIN and LIBERO benchmarks. I. Chunk and A. Chunk represent image chunk length and action chunk length, respectively}
\label{tab:chunk_ablation}
\end{table}

We first fix the image chunk length to 1, as adjacent image frames contain redundant information and predicting more images increases computational overhead, thereby reducing model efficiency. By varying the action chunk length, we observe that
\textbf{\textit{longer action sequences lead to degraded performance}}. We attribute this to error accumulation in action predictions, where mistakes in early actions compound over longer sequences. However, excessively short action chunk lengths also result in performance degradation, as the network learns less information from the dataset. 

On LIBERO, we obtain consistent results, with performance deteriorating as action chunk length increases. Based on these findings, we adopt action chunk lengths of 5 for CALVIN and 3 for LIBERO in our main experiments.

\subsection{Number of Chunks} 

To investigate the effect of the number of chunks, we fix the image chunk length to 1 and the action chunk length to 3, conducting comprehensive experiments with the number of chunks varying from 2 to 6. As illustrated in Figure~\ref{fig:number_of_chunks}, we observe the \textbf{same pattern} as in our experiments with image chunk lengths and action chunk lengths.

\textbf{\textit{A larger number of chunks require the model to predict more action steps, which leads to accumulation of prediction errors. And the decreasing rate becomes larger when the number of chunks increases. However, an excessively small number of chunks limits the model's ability to learn the relationship between world modeling and actions during training.}} 

According to our results, setting the number of chunks to 3 achieves the best performance.

We apply this finding to the configuration with image chunk length and action chunk length equal to 1 and 5, thereby achieving state-of-the-art results. This demonstrates

\textbf{\textit{Both the scalability and robustness of our baseline approach, validating that the optimal chunk configuration principles discovered through systematic experimentation can be effectively transferred to different parameter settings to achieve superior performance.}}

\section{Test-Time Training Ablation Study}

\subsection{TTT Details}
\label{app:ttt_details}

For CALVIN, since the test scenarios are relatively uniform within each environment, we do not need to perform TTT for every individual sample. Instead, we conduct adaptation by sampling only on the first sample's scenario. We select 34 text instructions that have appeared in the training set as guidance and perform random exploration for 60 frames. We then uniformly sample 6 times within these 60 frames for each task, which leads to $34\times6=204$ trajectories in total. Each trajectories have the same length as the pretraining dataset. Following~\cite{kojima2025loratttlowranktesttimetraining}, we use single-epoch single-step optimization with LoRA rank set to 128, employing the AdamW optimizer with a learning rate of 0.005 and weight decay of 0.01.

For LIBERO, the evaluation involves 10 different test scenarios in the LIBERO-10 dataset, necessitating a different TTT strategy. We perform scene-level TTT sampling in each individual test scenario to adapt to the specific environmental conditions. Similar to CALVIN,  we uniformly sample for 6 times with 6 different random seed for 60 frames exploration, which generates $6 \times 6$ samples per scenario. Also we set LoRA rank to 64 and  employ single-epoch single-step optimization using the AdamW optimizer with a learning rate of 0.005 and weight decay of 0.01.

\subsection{LoRA vs. Full Fine-tuning} 

We conducted a comparative study between LoRA and Full Fine-Tuning (FFT) on the CALVIN dataset. In our experiments, we set the LoRA rank to 128. Both LoRA fine-tuning and FFT used identical hyperparameters: learning rate $lr = 0.0005$, Adam optimizer, and weight decay of $0.01$.

\begin{table}[h]
\centering
\begin{tabular}{lc}
\hline
Method & Avg. Len. \\
\hline
WorldAgen & 3.87 \\
WorldAgen $+$ LoRA & \textbf{3.93} \\
WorldAgen $+$ FFT & 3.85 \\
\hline
\end{tabular}
\caption{Comparison of test-time training with LoRA and FFT on CALVIN dataset. Avg. Len. represents the average sequence length.}
\label{tab:lora_vs_fft}
\end{table}

As shown in Table \ref{tab:lora_vs_fft}, our results reveal that FFT does not improve model performance and actually leads to a decline in performance. We attribute this phenomenon to the fact that

\textbf{\textit{FFT causes the network to overfit to the test scenarios, whereas LoRA enables the model to acquire scenario-specific features while preserving the original scene perception capabilities. Therefore, LoRA demonstrates superior performance by striking a better balance between adaptation to new scenarios and retention of pretrained knowledge.}}

\subsection{Scalability and Robustness} To investigate the scalability and robustness of our method, we select three parameters crucial to TTT: number of chunks ($N$), image chunk length ($n$), and action chunk length ($m$). We conducted TTT experiments across different combinations of $N$, $n$, and $m$ values.

We choose three representative settings with different configurations to demonstrate the generalizability of our approach across various parameter combinations.

\begin{table}[h]
\centering
\begin{tabular}{ccccc|c}
\hline
$N$ & $n$ & $m$ & LoRA Rank & Data Size & \begin{tabular}{c} Before TTT \\ After TTT \end{tabular} \\
\hline
3 & 1 & 3 & 64 & 170 & \begin{tabular}{c} 3.89 \\ \textbf{3.90} \end{tabular} \\
\hline
5 & 1 & 3 & 256 & 204 & \begin{tabular}{c} 3.50 \\ \textbf{3.67} \end{tabular} \\
\hline
3 & 1 & 5 & 128 & 204 & \begin{tabular}{c} 3.87 \\ \textbf{3.93} \end{tabular} \\
\hline
\end{tabular}
\caption{TTT experiments under different number of chunks ($N$), image chunk length ($n$), and action chunk length ($m$) configurations.}
\label{tab:ttt_scalability}
\end{table}
As demonstrated in the Table \ref{tab:ttt_scalability}, TTT consistently improves performance across different settings using only around 200 samples. This validates 

\textbf{\textit{The scalability and robustness of our TTT approach and it can effectively adapt to various architectural configurations while maintaining consistent improvement patterns.}}

\begin{figure}[h]
\centering
\includegraphics[width=0.48\textwidth]{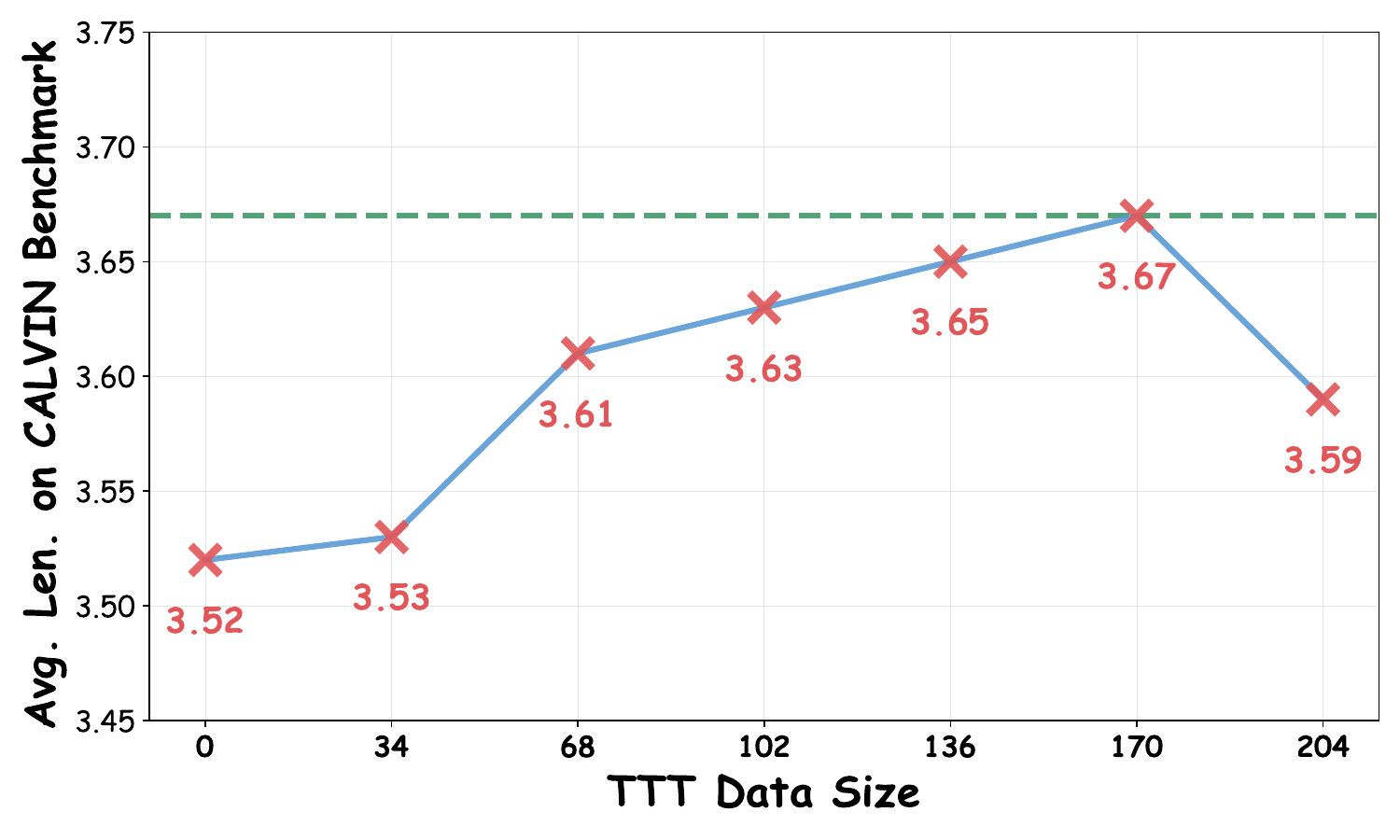}
\caption{This figure shows that as the amount of data increases, the TTT effect gradually increases.}
\label{fig:ttt_data_size}
\end{figure}

\subsection{Test-Time Training Data Size}
Furthermore, during the grid search process for the experimental configuration with $n=1$, $m=3$, $N=5$, as illustrated in the Figure \ref{fig:ttt_data_size}, we discovered that TTT exhibits linear performance improvement with small amounts of data. However, when the data volume becomes excessive, we observe a phenomenon where TTT performance degrades. These experiments were conducted with LoRA rank = 256 and learning rate = 0.0005.

We attribute this phenomenon to the fact that excessive training samples cause LoRA to overfit on the world modeling component, making the model more biased towards predicting world observations rather than executing correct actions. This finding highlights

\textbf{\textit{The importance of carefully balancing the amount of test-time training data to achieve optimal performance without compromising the model's action execution capabilities.}}

\subsection{Comparison Between Backbones}

\begin{table}[h]
\centering
\begin{tabular}{lcc}
\toprule
Backbone & CALVIN & LIBERO-10 \\
\midrule
GPT2   & 3.32 & 75.0 \\
Qwen3  & 3.87 & 75.5 \\
\bottomrule
\end{tabular} 
\caption{Performance comparison on different backbones.}
\label{tab:backbone_comparison}
\end{table}

We incorporate the GPT-2 \cite{radford2019language} backbone to examine the performance gap across different language backbones. Compared with GPT-2, Qwen3 demonstrates stronger representation capacity and better alignment with multimodal instructions, leading to consistently higher performance, as shown in Table~\ref{tab:backbone_comparison}.

\subsection{LIBERO Subsets}

\begin{table}[h]
\centering

\begin{tabular}{cccc}
\toprule
Spatial & Object & Goal & LIBERO-10 \\
\midrule
93.0 & 79.5 & 91.0  & 79.0 \\
\bottomrule
\end{tabular} 
\caption{Performance across different subsets in LIBERO.}
\label{tab:libero_subsets}
\end{table}

We further include additional subsets of the LIBERO benchmark to comprehensively evaluate the performance of WorldAgen. The results demonstrate that our method maintains stable and consistent performance across various subsets, indicating strong generalization ability and robustness to task diversity, as shown in Table~\ref{tab:libero_subsets}.

\subsection{Comparison Visualization}

We visualize the comparison before and after applying TTT. As shown in the figure \ref{fig:demos}, the red regions represent the failed portions of the trajectory, while the green regions denote the successful units. The selected task sequence is \textit{rotate\_blue\_block\_right}, \textit{move\_slider\_right}, \textit{lift\_red\_block\_slider}, and \textit{place\_in\_slider}. It can be observed that before applying TTT, the robot only completed the \textit{rotate\_blue\_block\_right} task and then entered a halted state. \textbf{\textit{After incorporating TTT, however, the model demonstrates a significantly enhanced ability to model the environment, thereby successfully accomplishing a series of tasks.}}

\begin{table*}[t]
\centering
\caption{Summary of important notations used in the WorldAgen framework.}
\label{tab:notation}
\begin{tabularx}{\textwidth}{llX}
\hline
Notation & Symbol & Description \\
\hline

Action & $a_t$ &
Robot action at time step $t$ in the raw rollout. \\[4pt]

Observation & $o_t$ &
Environmental observation at time step $t$ in the raw rollout. \\[4pt]

Robot State & $s_t$ &
Robot proprioceptive state at time step $t$ in the raw rollout. \\[4pt]

Rollout Trajectory & $\tau$ &
A trajectory with $T$ time steps:
$\tau = \{(o_t, s_t, a_t)\}_{t=0}^{T-1}$. \\[4pt]

Task Instruction & $G_i$ &
Language instruction specifying the task in unit $i$. \\[6pt]

\hline

Action Chunk Length & $m$ &
Number of actions in each action chunk $A_i$. \\[4pt]

Observation / Robot State Chunk Length & $n$ &
Number of observations and robot proprioceptive states in each chunk $O_i$ and $S_i$. \\[6pt]

\hline

Trajectory Unit & $U_i$ &
The $i$-th trajectory unit:
$U_i = (G_i, O_i, S_i, A_i, A_p, O_p)$. \\[10pt]

Action Chunk & $A_i$ &
Action chunk in unit $i$:
\[
\begin{aligned}
A_i &= A_{\,n+(i-1)m \rightarrow n+im-1} \\
    &= \big[
       a_{\,n+(i-1)m},\,
       a_{\,n+(i-1)m+1},\,
       \ldots,\,
       a_{\,n+im-1}
       \big].
\end{aligned}
\] \\[12pt]

Observation Chunk & $O_i$ &
Observation chunk in unit $i$:
\[
\begin{aligned}
O_i &= O^{n}_{\,n+(i-1)m \rightarrow n+im-1} \\
    &= \big[
       o_{\,n+(i-1)m},\,
       o_{\,n+(i-1)m + \lfloor \tfrac{1}{n} m \rfloor},\,
       \ldots,\,
       o_{\,n+(i-1)m + \lfloor \tfrac{n-1}{n} m \rfloor}
       \big].
\end{aligned}
\] \\[12pt]

Robot State Chunk & $S_i$ &
Robot state chunk in unit $i$:
\[
\begin{aligned}
S_i &= S^{n}_{\,n+(i-1)m \rightarrow n+im-1} \\
    &= \big[
       s_{\,n+(i-1)m},\,
       s_{\,n+(i-1)m + \lfloor \tfrac{1}{n} m \rfloor},\,
       \ldots,\,
       s_{\,n+(i-1)m + \lfloor \tfrac{n-1}{n} m \rfloor}
       \big].
\end{aligned}
\] \\[12pt]

\hline

Action Placeholder & $A_p$ &
Placeholder action tokens. \\[4pt]

Observation Placeholder & $O_p$ &
Placeholder observation tokens. \\[6pt]

Predicted Action Chunk & $\hat{A}_i$ &
Predicted action chunk:
\[
\begin{aligned}
\hat{A}_i &= 
\big[
\hat{a}_{\,n+(i-1)m},\,
\hat{a}_{\,n+(i-1)m+1},\,
\ldots,\,
\hat{a}_{\,n+im-1}
\big].
\end{aligned}
\] \\[12pt]

Predicted Observation Chunk & $\hat{O}_i$ &
Predicted next observation chunk:
\[
\begin{aligned}
\hat{O}_i &= \hat{O}^{n}_{\,n+im \rightarrow n+(i+1)m-1} \\
          &= \big[
             \hat{o}_{\,n+im},\,
             \hat{o}_{\,n+im + \lfloor \tfrac{1}{n} m \rfloor},\,
             \ldots,\,
             \hat{o}_{\,n+im + \lfloor \tfrac{n-1}{n} m \rfloor}
             \big].
\end{aligned}
\] \\[12pt]

\hline
\end{tabularx}
\end{table*}

\end{document}